\documentclass[10pt]{amsart}
\usepackage{geometry}                
\usepackage{graphicx}
\usepackage{amssymb}
\usepackage{epstopdf}
\usepackage{latexsym}
\usepackage{amsmath}
\usepackage{gensymb}
\usepackage{hyperref}
\begin{document}


\title{A convolutional approach to reflection symmetry}
\author{Marcelo Cicconet, Vighnesh Birodkar, Mads Lund,\\Michael Werman, and Davi Geiger}

%
%

\begin{abstract}
We present a convolutional approach to reflection symmetry detection in 2D. Our model, built on the products of complex-valued wavelet convolutions, simplifies previous edge-based pairwise methods. Being  parameter-centered, as opposed to feature-centered, it has certain computational advantages when the object sizes are known a priori, as demonstrated in an ellipse detection application. The method outperforms the best-performing algorithm on the CVPR 2013 Symmetry Detection Competition Database in the single-symmetry case. Code and a new database for 2D symmetry detection is available.
\end{abstract}

%




\maketitle

\section{Introduction}

\begin{figure}[h]
\centering
\vspace{1cm}
\begin{minipage}[c]{0.3\linewidth}
	\centering
	\includegraphics[width=\linewidth]{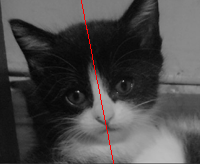}\\(a)
\end{minipage}\hspace{1cm}
\begin{minipage}[c]{0.3\linewidth}
	\centering
	\includegraphics[width=\linewidth]{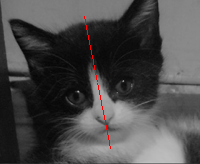}\\(b)
\end{minipage}\\ \vspace{0.3cm}
\begin{minipage}[c]{\linewidth}
	\centering
	\includegraphics[width=0.6\linewidth]{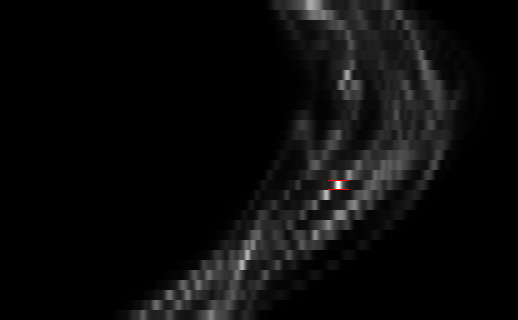}\\(c)
\end{minipage}\\ \vspace{0.3cm}
\begin{minipage}[c]{0.3\linewidth}
	\centering
	\includegraphics[width=\linewidth]{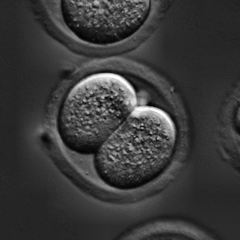}\\(d)
\end{minipage}\hspace{1cm}
\begin{minipage}[c]{0.3\linewidth}
	\centering
	\includegraphics[width=\linewidth]{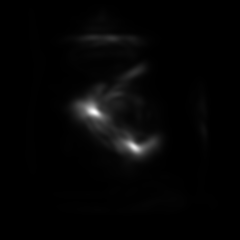}\\(e)
\end{minipage}\\ \vspace{0.3cm}
\vspace{0.5cm}
\caption{(a) Symmetry line. (b) Symmetry segment. (c) Accumulator space for symmetry line detection. Rows correspond to angle $\rho$ and columns to displacement $\delta$. The red mark corresponds to the line shown in (a). (d) Mouse embryo cells.
(e) Accumulator space for ellipse detection from (d).}
\label{fig:modes}
\end{figure}

In this paper we are primarily interested in (1) detecting the line of reflection (mirror) symmetry in 2D and, (2) given such a line, finding the straight \emph{segment} that divides the symmetric object into its mirror symmetric parts (Figure~\ref{fig:modes}~(a,b)). As a byproduct, we develop a coefficient (likelihood) of reflection symmetry, that can be used to locate near circular shapes (Figure~\ref{fig:modes}~(e)).

Symmetry detection algorithms can be classified as: feature-centered, or parameter-centered. Feature-centered; where image features (whatever they might be) are computed first, and the symmetry parameters are computed from them. Parameter-centered; where we start from the parameters, and then look for image features that support them.

To the best of our knowledge, all previous works fall in the first category. The approach presented here falls in the second. Our symmetry likelihood model is based on products of complex-valued wavelet convolutions. This parameter-centered approach has advantages when the object sizes are known a priori, as demonstrated in an ellipse detection application.

In addition, we release a new database for 2D symmetry detection, with  4+ times as many single symmetry images, and 2+ times as many multiple symmetry images as the most recently published database \cite{Liu2013}.

We compare our model against the state of the art in \cite{Liu2013}, outperforming it on the single-symmetry case.

Code: \href{https://github.com/cicconet/SymmetryAxes}{https://github.com/cicconet/SymmetryAxes}

Database: \href{http://symmetry.cs.nyu.edu}{http://symmetry.cs.nyu.edu}

Note: This paper is under consideration at Pattern Recognition Letters.

\section{Previous Work}

\begin{figure}[p]
\centering
\vspace{1cm}
\includegraphics[width=0.7\linewidth]{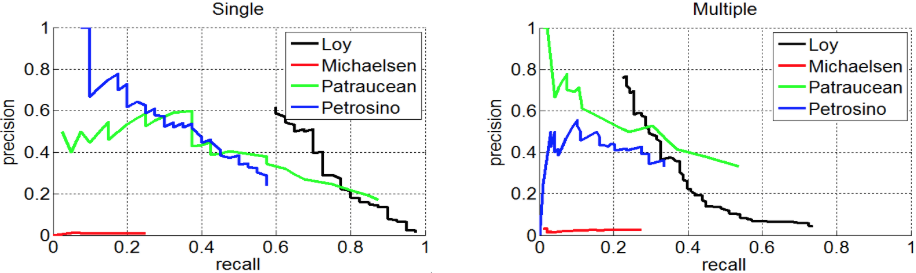}
\caption{Results of the CVPR 2013 Symmetry Detection from Real World Images Competition, as shown in \cite{Liu2013}. Compare to Figure~\ref{fig:results}.}
\label{fig:prevwork}
\end{figure}

Four different methods for mirror symmetry segment detection were compared in \cite{Liu2013} -- CVPR 2013 Competition: \cite{Kondra2013, Patraucean2013, Michaelsen2013, Loy2006}. Figure~\ref{fig:prevwork}, taken from the competition report, shows their performance.

In \cite{Kondra2013}, the algorithm operates in three steps: (1) correlation measures (using the SIFT descriptor) are computed along discrete directions; (2) symmetrical regions are identified by looking for matches in the directions characterized by maximum correlations; steps (1) and (2) are performed at different scales.

\cite{Patraucean2013} runs in two steps: (1) candidates for mirror-symmetric patches are identified using a Hough-like voting scheme; (2) candidates are validated using a principled statistical procedure inspired from a contrario theory.

In \cite{Michaelsen2013}, a combinatorial technique from Gestalt Algebra is used on top of SIFT descriptors.

Also based on SIFT is the work of \cite{Loy2006}. These features are grouped into ``symmetric constellations,'' by a voting scheme, and the dominant symmetries present in the image emerge as local maxima.

Our method relates to \cite{Loy2006} and \cite{Patraucean2013} in the sense that we also use a voting scheme. But we use simple ``derivatives'' at different directions (given by wavelet filtering), instead of the complex SIFT descriptor.

This work can be seen as an extension of our previous work \cite{Cicconet2013}, where we accumulate wavelet responses for circle detection. Here, however, we add \emph{products} of wavelet convolutions, instead of adding convolutions directly. As we will show in the applications, detection of near-circular shapes can be done as a particular case of the 
symmetry detection framework -- as a circle is mirror symmetric with respect to every diagonal.

\section{Symmetry Line}

\begin{figure}[p]
\centering
\vspace{1cm}
\includegraphics[width=0.3\linewidth]{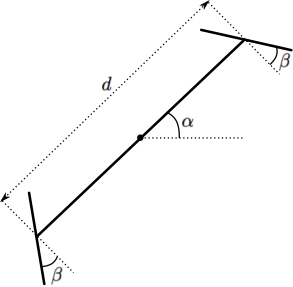}
\caption{Stencil. Symmetry likelihood is computed by accumulating the products of wavelet
outputs at the ends of the stencil while it orbits around the image. The
range covered by each of the parameters depends on the application and dataset.}
\label{fig:stencil}
\end{figure}

The core of our method is the accumulation of
the products of convolutions with symmetric,
complex-valued wavelet kernels.
We refer to the orbiting
pair of symmetric wavelets as a \emph{stencil} (Figure~\ref{fig:stencil}).

A stencil has three parameters: the angle $\alpha$ of
the line $l$ connecting the wavelets centers, the
angle $\beta$ of the wavelets -- measured with respect
to the perpendicular to $l$ at the point where the wavelets are
mounted, and the distance $d$ between wavelet centers.
We sometimes refer to $\alpha$ as the \emph{outer} angle,
$\beta$ as the \emph{inner} angle, and $d$ as the \emph{inner} distance.

\begin{figure}[p]
\centering
\vspace{1cm}
\includegraphics[width=0.2\linewidth]{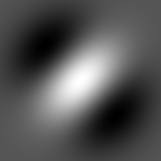}\hspace{0.5cm}
\includegraphics[width=0.2\linewidth]{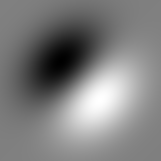}
\caption{Real and imaginary parts of wavelet kernels of angle $\pi/4$. The real component is better at detecting ridge-type edges, while the imaginary part fires more strongly at step-type edges.}
\label{fig:wavelets}
\end{figure}

We use Morlet wavelets (\cite{Bruna2013}). Their shape is illustrated
in Figure~\ref{fig:wavelets}. These kernels are
essentially the product of a Gaussian-like envelope
with a complex, oscillating carrier, normalized
to have mean zero and magnitude one.
We use complex wavelets, rather than either
of its real-valued components, as they provide
a better trade-off in capturing both ridge-type and step-type edges
(the real part would be more suited to the former, and the imaginary
part to the latter).

Let $p$ be a point (pixel location) in an image $I$.
Let $s = s_p(\alpha,\beta,d)$ be a stencil at a fixed configuration $\alpha,\beta,d$ centered at $p$. Let $v_s$ be the wavelet
kernel at the extreme $p_v = p+\frac{d}{2}(\cos(\alpha),\sin(\alpha))$ of the stencil, and $w_s$ the wavelet kernel at the opposite end, i.e., at $p_w = p-\frac{d}{2}(\cos(\alpha),\sin(\alpha))$.
Let $I_v = I * v_s$ and $I_w = I * w_s$, where \, $*$ \, is the convolution operator. The coefficient of mirror
symmetry for the stencil $s$ at the point $p$ is defined as

\begin{equation}
c^s_p = I_v(p_v)\cdot \overline{ I_w(p_w)}\text{ ,}
\end{equation}

\noindent
where \, $\bar{}$ \, is the complex-conjugate operator. The aggregate coefficient of mirror symmetry at $p$ is computed by integrating over all stencils centered at $p$:

\begin{equation}
c_p = \left| \sum_s (c^s_p) \right| ^n\text{ ,}
\end{equation}

\noindent
where the exponent $n$ controls the sharpness of the peaks in the accumulator space. Typical values are $n = 1$ and $n = 2$ (the optimal value depends on the dataset and problem).
The answer to the question of what ``all stencils'' means is determined empirically. In the applications reported here, $\alpha \in \{0,\frac{\pi}{n_\alpha},2\frac{\pi}{n_\alpha},...,\pi\}$, with $n_\alpha = 32$; $\beta \in \{-\frac{\pi}{4},0,\frac{\pi}{4}\}$ for symmetry detection, and $\beta = 0$ for ellipse detection; $d \in \{d_1,...,d_N\}$ varies in equally spaced samples (every 2 to 5 pixels), between close to $0$ and a factor of the size of the image for symmetry detection, or around the expected diameters of the ellipses in ellipse detection. In other words, $\alpha$ covers a dense-enough sample of the interval $[0,\pi]$, $\beta$ covers a few directions around $0$, and $d$ depends on the expected size of the symmetric shape.

In summary, given a point $p$, a set of stencils $s(\alpha,\beta,d)$ accumulates the
evidence $c_p$ that $p$ lies in the line of symmetry of an object in the image. On the other hand, every triad $(\alpha,\beta,d)$ captures evidence that there is a symmetric shape with a symmetric line perpendicular to the corresponding stencil -- the evidence is given by the product of the wavelet outputs at the ends of the stencil. This is what we mean by a parameter-centered approach: we start from a triad $(\alpha,\beta,d)$ and ask: is there evidence in the image supporting the existence of a symmetry line perpendicular to the stencil $s(\alpha,\beta,d)$?



When the symmetry likelihood $c_p$ is \emph{stored} at $p$,
the resulting accumulator space (which coincides with the image space) can be used to locate
the centers of ellipses and circles in images (Figure~\ref{fig:modes}~(d,e)).

To find symmetry lines, we have to accumulate votes in
a parameter space of lines. Similarly to the Hough-transform
for line detection (\cite{Duda1972}), we use the space $\rho,\delta$ of
the angles and displacements of the line perpendicular to the stencil $s(\alpha,\beta,d)$ centered at $p$ (Figure~\ref{fig:modes}~(c)). Such line is given by

$$\rho = \alpha+\frac{\pi}{2}\text{ ,} \hspace{0.5cm}
\delta = p_x \cdot \cos \alpha + p_y \cdot \sin \alpha\text{ ,}$$

\noindent
where $p_x,p_y$ are the components of $p$.

\begin{figure}[p]
\centering
\vspace{1cm}
\begin{minipage}[c]{\linewidth}
	\centering
	\includegraphics[width=\linewidth]{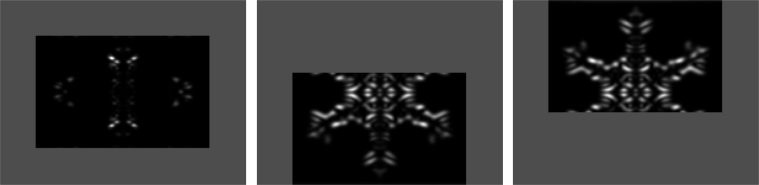}\\(a)
\end{minipage}
\\ \vspace{0.3cm}
\begin{minipage}[c]{\linewidth}
	\centering
	\includegraphics[width=\linewidth]{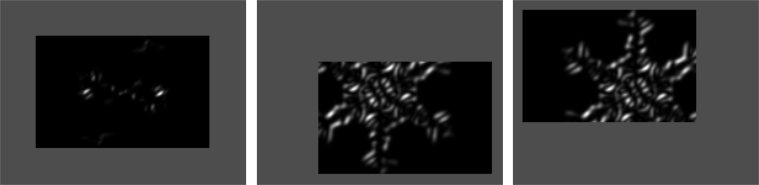}\\(b)
\end{minipage}
\\ \vspace{0.3cm}
\begin{minipage}[c]{\linewidth}
	\centering
	\includegraphics[width=\linewidth]{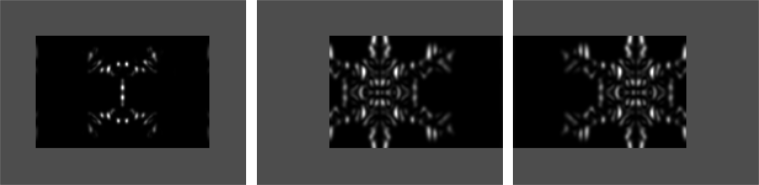}\\(c)
\end{minipage}
\\ \vspace{0.3cm}
\begin{minipage}[c]{\linewidth}
	\centering
	\includegraphics[width=\linewidth]{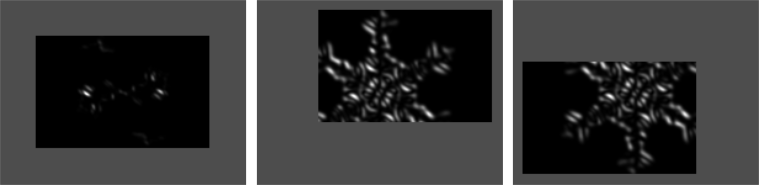}\\(d)
\end{minipage}
\caption{(a)/(b)/(c)/(d) Illustration of the block-based computation in the intermediate step of the likelihood model, in which the stencil is rotated at outer angles 0$\degree$/45$\degree$/90$\degree$/135$\degree$, respectively. The cropped image on the left is the product of the cropped image in the middle with the cropped image on the right.}
\label{fig:offsets}
\end{figure}

The implementation of the method as described
above is computationally very expensive, for it requires looping through all possible parameters $p,\alpha,\beta,d$. Notice,
however, that for fixed $(\alpha,\beta,d)$, the kernels at the ends of a stencil are invariant, and therefore the convolutions can be computed beforehand, and their product added in blocks at proper offset locations. This is illustrated in Figure~\ref{fig:offsets}.
  
This step can also be used to fine-tune the likelihoods of symmetry lines computed in the previous step, by sorting them according to the likelihood computed via robust statistics (more details in the next subsection). In certain applications/datasets this provides better results.

\section{Symmetry Segment}

\begin{figure*}[t!]
\centering
\vspace{1cm}
\begin{minipage}[c]{0.25\linewidth}
	\centering
	\includegraphics[height=3.5cm]{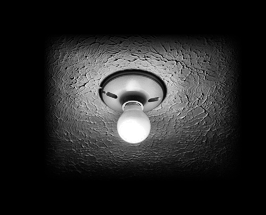}\\(a)
\end{minipage}\hspace{0.5cm}
\begin{minipage}[c]{0.1\linewidth}
	\centering
	\includegraphics[height=3.5cm]{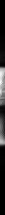}\\(b)
\end{minipage}\hspace{0.1cm}
\begin{minipage}[c]{0.2\linewidth}
	\centering
	\includegraphics[height=3.5cm]{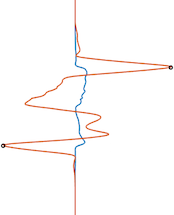}\\(c)
\end{minipage}\hspace{0.5cm}
\begin{minipage}[c]{0.25\linewidth}
	\centering
	\includegraphics[height=3.5cm]{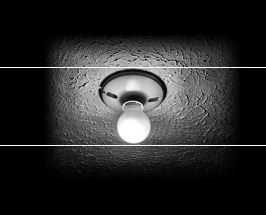}\\(d)
\end{minipage}
\caption{Detection of end-points, given a symmetry line. (a) Input example. (b) Five highest symmetry votes from stencils perpendicular to the symmetry line. (c) In blue, sum of the columns in (b); in red, convolution of the blue line with a Haar wavelet; this allows for easy peak detection -- black dots. (d) Display of end points as white lines, characterizing the line segment capturing the lamp.}
\label{fig:endpoints}
\end{figure*}

The end points of the symmetry line are computed as illustrated in Figure~\ref{fig:endpoints}.
Let $l$ be a symmetry line found as described above,
and $p$ a point in such line.
For all stencils $s$ centered in $p$ and perpendicular
to $l$ we accumulate the five votes that have highest symmetry across $\beta$ and $d$, i.e., along the line $l$ we create a robust histogram of symmetry likelihoods (Figure~\ref{fig:endpoints}~(b)). We then pose the problem of detecting the end points of the segment as that of finding a pair of points along $l$ where the histogram changes the most. These points are computed as the extreme/peaks of the convolution of the histogram with a Haar wavelet ((Figure~\ref{fig:endpoints}~(c)). The size of the wavelet is determined empirically (the value of 20 is used in the experiments reported here).

\section{Experiments}
\label{sec:experiments}

Our method was  tested on the   CVPR 2013 Symmetry Detection from Real World Images Competition database (\cite{Liu2013}), as well as our own database.
The distribution of the number of images is the following:

\begin{table}[h]
\begin{tabular}{c c c}
   & CVPR 2013 dataset & our dataset \\
  single symmetry axis & 40 & 176 \\
  multiple symmetry axes & 30 & 63 \\
\end{tabular}
\end{table}

Both datasets provide the end points of the symmetric shape. All the images were resized, keeping the aspect ratio, so that the maximum dimension is $200$ pixels. Our dataset will be made public with this paper.

\subsection{Metric}

In order to compare our approach to previous methods, we use the same metric for accuracy as in \cite{Liu2013}. More precisely, let $\phi_G$, $c_G$, and $L_G$ be the angle, center point, and segment length, respectively, of a ground truth \emph{segment} $t_G$ in the database\footnote{By \emph{segment} we mean a line delimited by end points. The unlimited line is referred to simply as \emph{line}. The line corresponding to $t_G$ is denoted as $l_G$.}. The algorithm proposes several segments $t_D$ as solutions, with parameters $\phi_D$, $c_D$, and $L_D$ (analogous to $\phi_G$, $c_G$, and $L_G$). Let $\Delta_{G,D}$ be the distance between the centers $c_G$ and $c_D$. Segment $t_D$ is considered to correctly correspond to the segment $t_G$ (that is, a true positive, or TP) if

$$|\phi_G-\phi_D| < 10^\circ \text{ end } \Delta_{G,D} < \frac{1}{5} min\{L_G, L_D\}\text{ .}$$

\noindent
Notice how the metric for segment detection gives no incentive for the proper location of the \emph{end points}, only the \emph{center}. Indeed, for segment detection, one could devise an algorithm that only looks for the center of the line $c_D$, and outputs $L_D = \infty$, so as to maximize the probability of $\Delta_{D,G} < \frac{1}{5} min\{L_G, L_D\}$ being satisfied. Regardless, our algorithm does output a finite length and is evaluated that way.

Accuracy for \emph{line} detection is measured in a similar fashion. Let now $\Delta^l_{G,D}$ be the distance between the \emph{point} $c_G$ (center of ground truth) and the \emph{line} $l_D$. Line $l_D$ is considered to correctly correspond to segment $t_G$ if

$$|\phi_G-\phi_D| < 10^\circ \text{ end } \Delta^l_{G,D} < \frac{1}{5}L_G\text{ .}$$

Notice that the distance between a line $l_D$ and a segment  $t_G$ is never worse than the distance between a segment along a line, $t_D$ ,  and the same segment $t_G$. This is because to estimate the distance between a line and a center, the center chosen along the line $l_D$ is the closest one to the center $c_G$ of the segment $t_G$.   Thus, our line detection will always perform better than when adding to it the automatic estimation of the center of the segment. The reason we study line detection is to help us understand how much of the inaccuracy of our method is due to the line detection and how much is due to the center detection (segment detection).

A false negative (FN) consists of ground truth segment for which none of the lines proposed by the method satisfies the above condition. A false positive (FP) consists of a proposed line which is not ``close'' (as just defined) to any of the ground truth lines. Proposed lines that are close as per referred metric are counted only once, either for true positive or false positive measurement.  Precision and recall are defined as:

$$\text{precision}=\frac{TP}{TP+FP},\text{ recall} = \frac{TP}{TP+FN}\text{ .}$$

\subsection{Results}

\begin{figure*}[p]
\centering
\vspace{1cm}
\begin{minipage}[c]{0.49\linewidth}
\centering
\includegraphics[width=\linewidth]{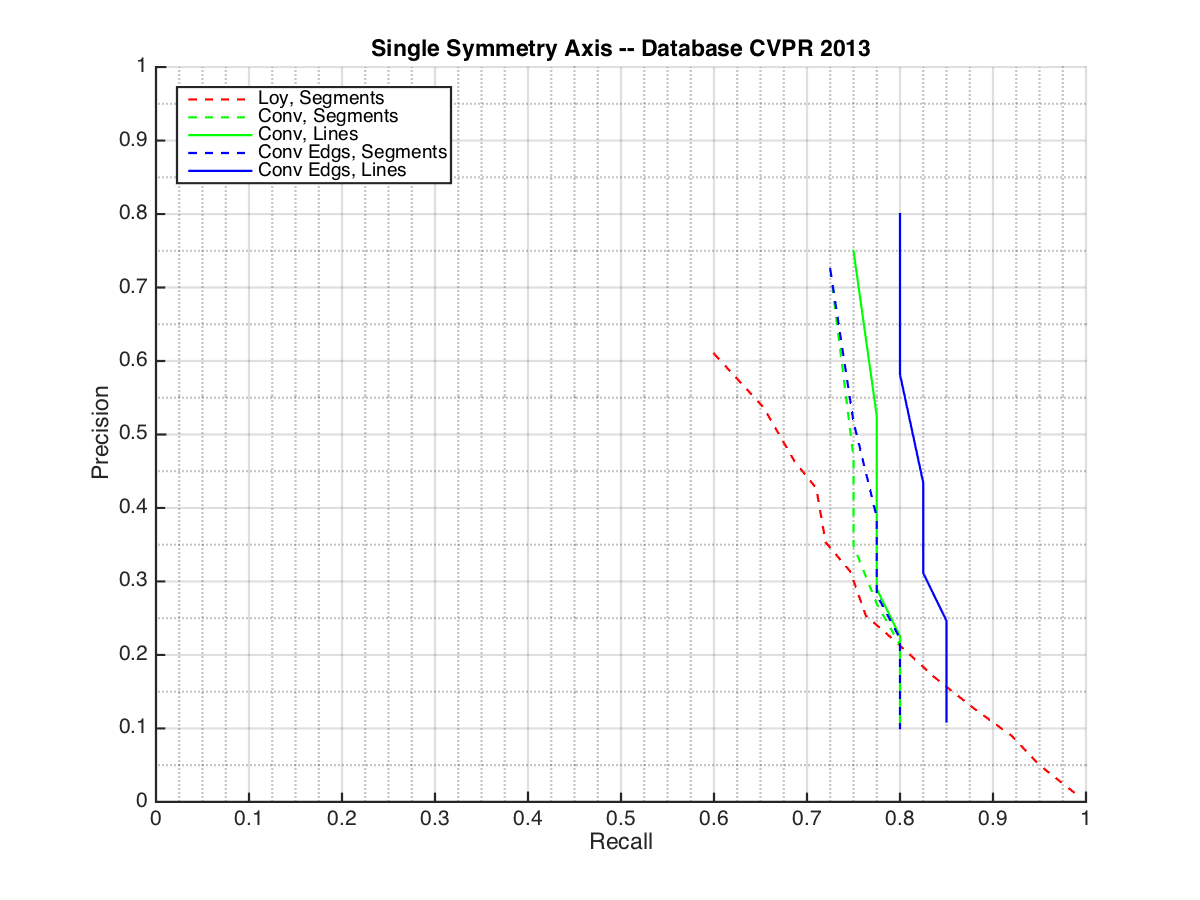}\\ \vspace{-0.3cm}(a)
\end{minipage}
\hfill
\begin{minipage}[c]{0.49\linewidth}
\centering
\includegraphics[width=\linewidth]{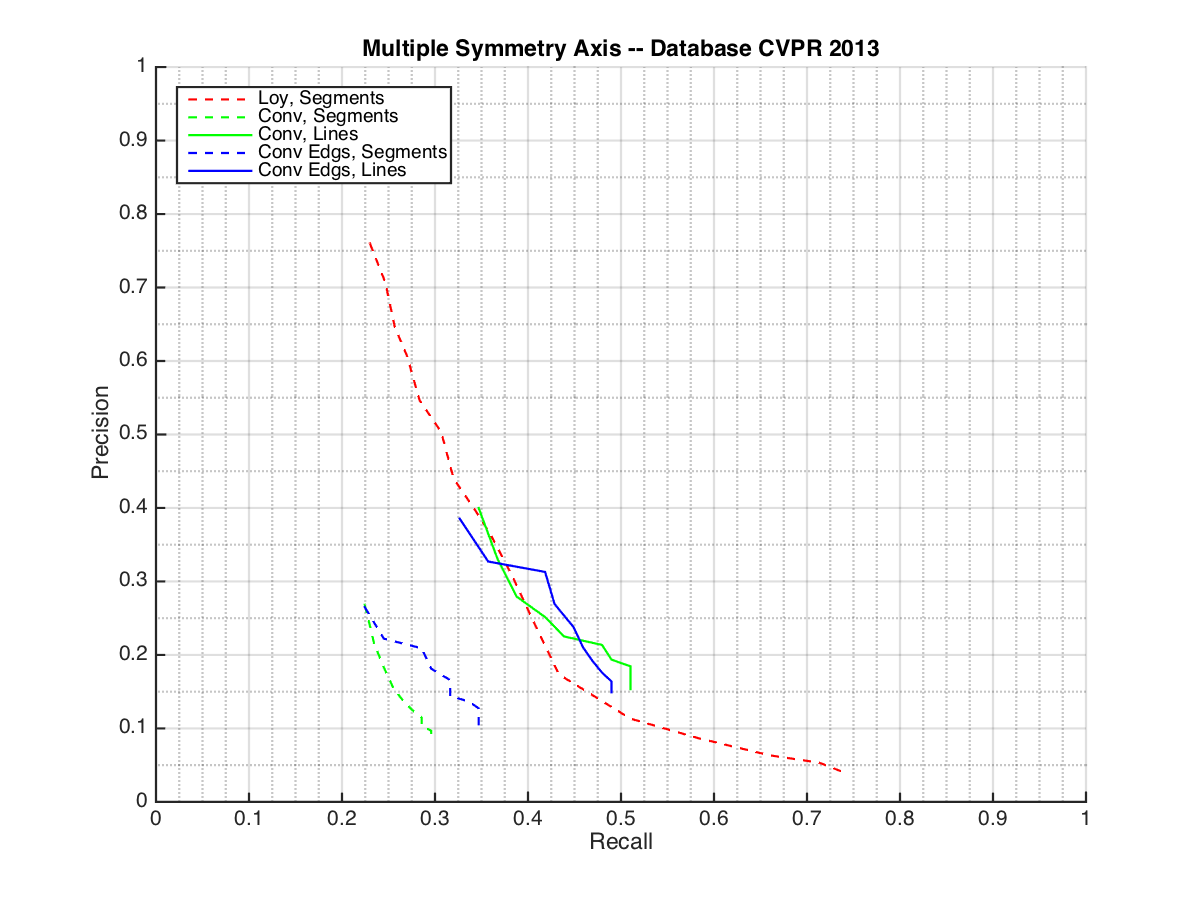}\\ \vspace{-0.3cm}(b)
\end{minipage}\\ \vspace{0.3cm}
\begin{minipage}[c]{0.49\linewidth}
\centering
\includegraphics[width=\linewidth]{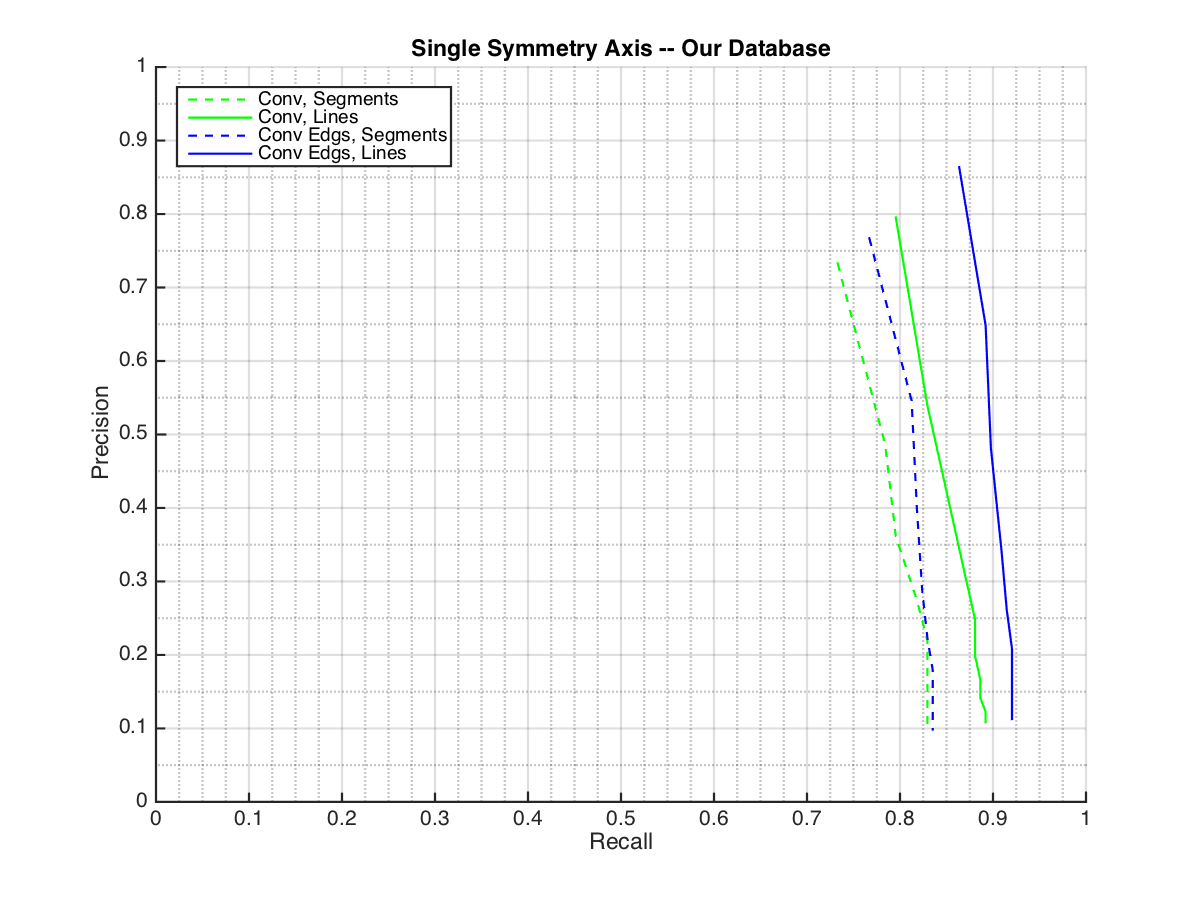}\\ \vspace{-0.3cm}(c)
\end{minipage}
\hfill
\begin{minipage}[c]{0.49\linewidth}
\centering
\includegraphics[width=\linewidth]{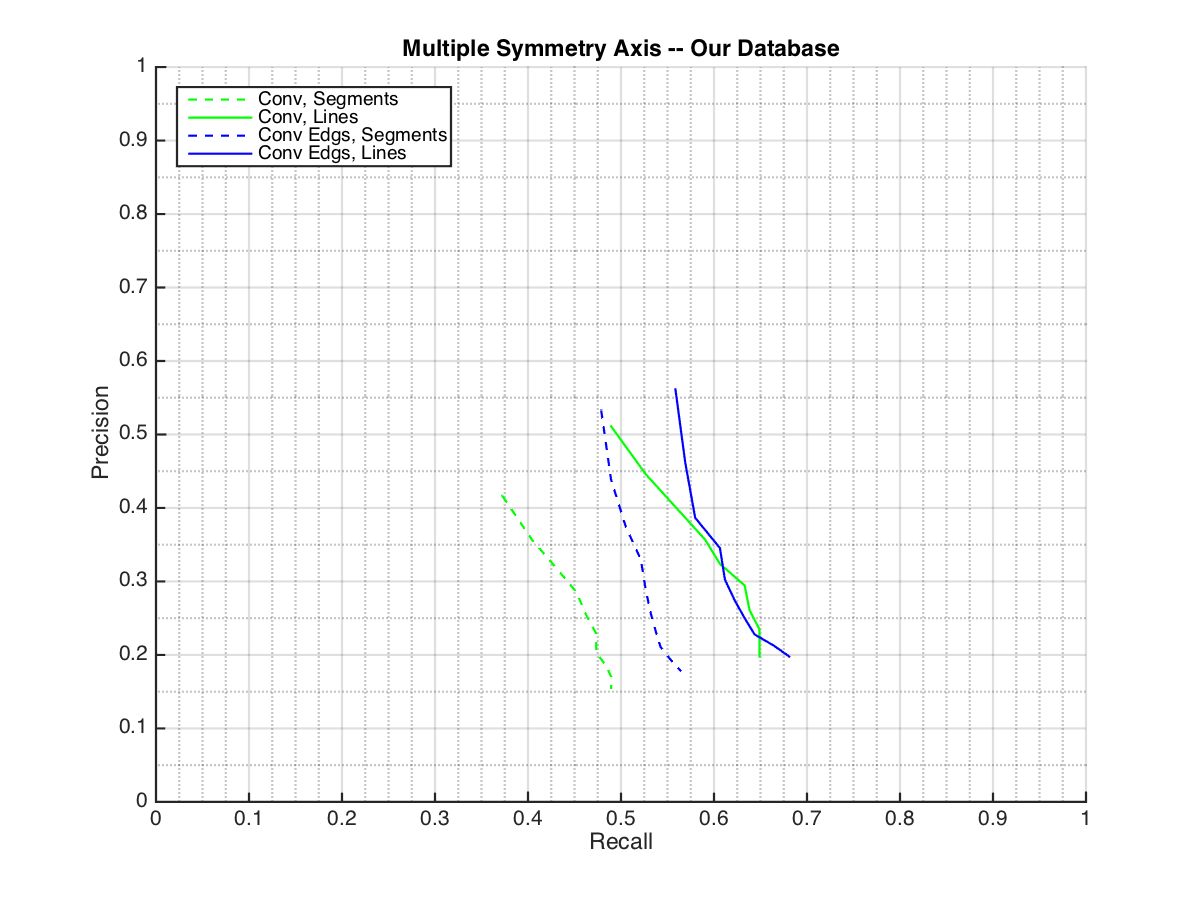}\\ \vspace{-0.3cm}(d)
\end{minipage}\vspace{0.5cm}
\caption{Performance results in terms of precision/recall curves. (a) Single symmetry axis detection on the CVPR 2013 database. (b) Multiple symmetry axes detection on the CVPR 2013 database. (c) Single symmetry axis detection in our database. (d) Multiple symmetry axes detection in our database. In the legends, ``Conv'' means our convolutional method as originally described, while ``Conv Edgs'' means the method applied on edge maps from the images. The latter is slightly better. Plots (a) and (b) display the results of the best-performing method in \cite{Liu2013}, namely \cite{Loy2006}. We  outperform \cite{Loy2006} in the single-symmetry case, but not for multiple symmetries -- some problems with the data is discussed in the text.}
\label{fig:results}
\end{figure*}

Our results are presented in Figure~\ref{fig:results}. Since our method essentially looks for symmetric edges, we expected that it would perform better when edges are  enhanced. Therefore, besides the basic method described previously, we also measured performance when applying the model on edge maps computed from the images.
We used the Ultrametric Contour Map algorithm, a general purpose state-of-the-art edge detector (\cite{Arbelaez2011}).

\begin{figure*}[t!]
\centering
\vspace{1cm}
\begin{minipage}[c]{0.23\linewidth}
	\centering
	\includegraphics[width=\linewidth]{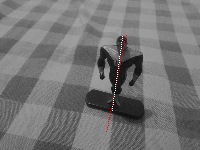}
\end{minipage}\hfill
\begin{minipage}[c]{0.23\linewidth}
	\centering
	\includegraphics[width=\linewidth]{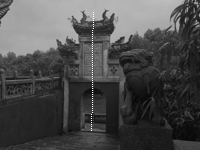}
\end{minipage}\hfill
\begin{minipage}[c]{0.23\linewidth}
	\centering
	\includegraphics[width=\linewidth]{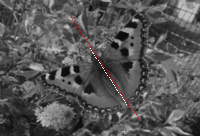}
\end{minipage}\hfill
\begin{minipage}[c]{0.23\linewidth}
	\centering
	\includegraphics[width=\linewidth]{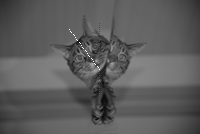}
\end{minipage}\\ \vspace{0.3cm}
\begin{minipage}[c]{0.23\linewidth}
	\centering
	\includegraphics[width=\linewidth]{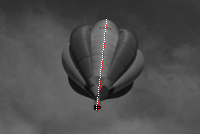}
\end{minipage}\hfill
\begin{minipage}[c]{0.23\linewidth}
	\centering
	\includegraphics[width=\linewidth]{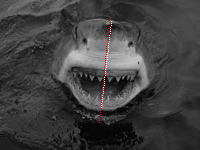}
\end{minipage}\hfill
\begin{minipage}[c]{0.23\linewidth}
	\centering
	\includegraphics[width=\linewidth]{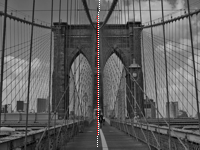}
\end{minipage}\hfill
\begin{minipage}[c]{0.23\linewidth}
	\centering
	\includegraphics[width=\linewidth]{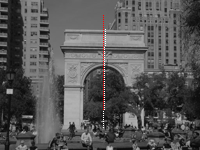}
\end{minipage}
\caption{Some examples of outputs from our algorithm on our dataset (in red), along with the ground truth (in white).}
\label{fig:outputexamples}
\end{figure*}

As expected, performance is better on edge maps, and when the distance from extracted lines $l_D$ to the ground truth segments $t_G$ is used instead of the distance between $t_D$ to $t_G$. Some examples of outputs (best guess, using the convolutional model directly on images) are shown in Figure~\ref{fig:outputexamples}. 

As per Figure~\ref{fig:results}~(a), both versions outperform the state-the-art \cite{Loy2006} in single symmetry \emph{segment} detection on the CVPR 2013 database (see Figure~\ref{fig:results}~(a)). Line detection for the multiple-symmetry case is competitive, but the center detection is quite weak in this case (worsening the results significantly from line detection). Future work could attempt to improve the center detection along the symmetry line; for example by identifying the clusters of data that vote for the line and outputting the center from the largest cluster only.

Moreover, as we discuss next, there are some problems with the ground truth labeling in the multiple symmetry case, which may explain the poor performance of center detection.

\subsection{Comments on the Database}
\label{sub:comments}

Looking at the labeling of the CVPR 2013 database, we see that, in particular for the multiple symmetries case, there are images with inaccurate and inconsistent  ground truth. As a matter of fact, the group that organized the corresponding CVPR 2013 workshop recently published a paper that recognizes the issue (\cite{Funk2016}).

\begin{figure*}[t!]
\centering
\vspace{1cm}
\begin{minipage}[c]{0.32\linewidth}
	\centering
	\includegraphics[height=3cm]{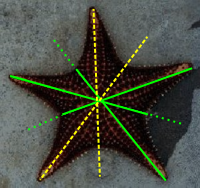}\\(a)
\end{minipage}\hfill
\begin{minipage}[c]{0.32\linewidth}
	\centering
	\includegraphics[height=3cm]{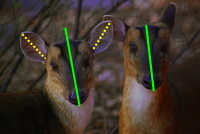}\\(b)
\end{minipage}\hfill
\begin{minipage}[c]{0.32\linewidth}
	\centering
	\includegraphics[height=3cm]{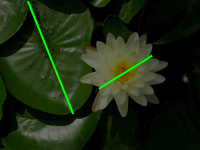}\\(c)
\end{minipage}
\caption{Problems with CVPR 2013's ground truth for multiple symmetry axis. Green solid segments are the ground truth. Green dashed segment extensions represent ground truth we believe are more consistent, since the dataset often has segments beyond the object boundary if there is symmetry support. Yellow dashed segments represent segments not included in the ground truth. All these problems contribute to ground truth centers that are inaccurate and for false positive, as well as angle errors that should not occur when evaluating an algorithm.}
\label{fig:problems}
\end{figure*}

We discuss here some examples. In Figure \ref{fig:problems}~(a) the ground truth segments are represented by the highlighted  green solid line segments. It can be observed that they terminate at the boundary of the object, but to be consistent with other ones in the data set, like Figure \ref{fig:problems}~(b)  the symmetry should extend beyond that, which is highlighted with green dotted lines. This distinction significantly changes the center of the segments and therefore the performance of any algorithm on it, according to the current metric. Moreover, two additional symmetry line segments are missing, highlighted here in yellow dashed lines. In Figure \ref{fig:problems}~(b), we also notice the lack of identified ground truth symmetry segments along the ears of the deer, now marked by yellow dashed line. In Figure \ref{fig:problems}~(c), even though both segments are acceptable as ground truth, there are many others, which are rotations of the labeled one with respect to the center of the flower. An algorithm that outputs any of them should not be penalized for the angle error.

\begin{figure*}[p]
\centering
\vspace{1cm}
\begin{minipage}[c]{0.65\linewidth}
	\centering
	\includegraphics[width=\linewidth]{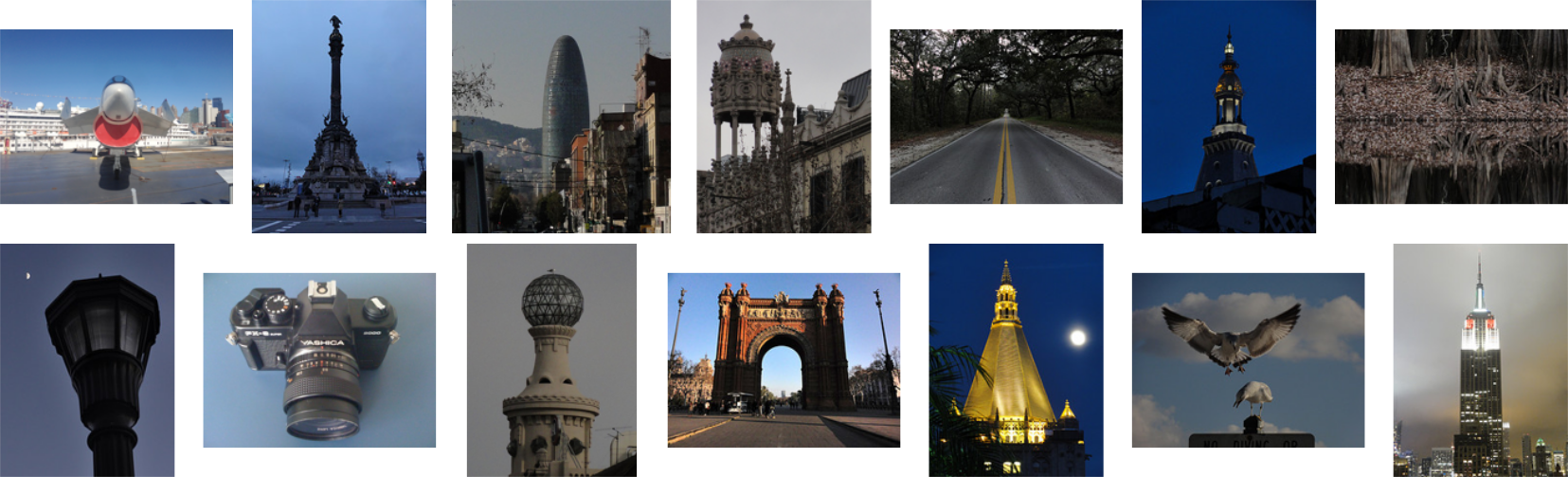}\\(a)
\end{minipage}\\
\begin{minipage}[c]{0.65\linewidth}
	\centering
	\includegraphics[width=\linewidth]{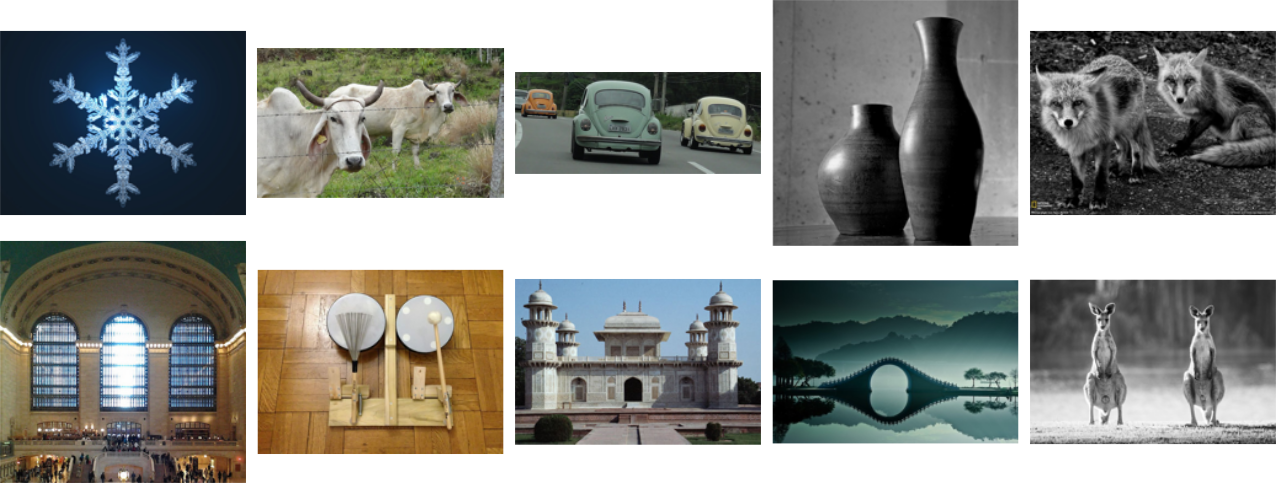}\\(b)
\end{minipage}
\caption{Sample images from our database. (a) Single symmetry. (b) Multiple symmetries. We gathered 176 images with single symmetry axis, and 63 with multiple axes. For comparison, the CVPR 2013 Competition database contains 40 and 30 images, respectively.}
\label{fig:newdatabase}
\end{figure*}

\section{Ellipse Detection}

A possible limitation of the convolutional approach
is that the stencil will often -- if not most of the time -- have at least one wavelet on top of an area where no edges exist. This is indeed a suboptimal approach for cases where there aren't many strong edges in the image.

On the other hand, existing methods start from edges, or other image features, and look at their pairs. In an image with many edges, computational cost can therefore be quite high.

If we know the properties of the objects, it is more efficient to optimize a parameter-centered method than a feature-centered one, because in the latter we have to loop through all features to compute the parameters in order to see if they are useful. One application in which this trade-off is apparent is in ellipse detection. To find the center of an ellipse using edges, one has to look not simply at pairs of tangents, but at triplets of tangents \cite{Mclaughlin98}, or, optionally, pairs of pairs of tangents (\cite{Cicconet2014ICIP1}).

Notice that if we restrict $\beta = 0$ and $d = 2r$, our method resembles the Hough Transform to detect circles of radius $r$.
Now, if we allow $d$ to vary around $2r$, we can, in principle, detect near-circular shapes of size $2r$ (in particular, ellipses with axes length around $2r$).

We did one such test on a dataset of mouse-embryo cells where images contain precisely two cells. One sample is shown in Figure~\ref{fig:modes}~(d), with the corresponding ellipse-center likelihood shown in Figure~\ref{fig:modes}~(e). The dataset is  from \cite{Cicconet2014ICIP1}, and contains 421 images.

We compared our algorithm to the feature-centered method in \cite{Cicconet2014ICIP1} (called Ellipses from Triangles, or EFT), both in terms of accuracy and computational time.
The average cell radii for the dataset was about 40 pixels, so we set the range of $d$ in $\{35,40,45\}$, and equivalent search range for the EFT.
The average time per image for the EFT was about 87.5 seconds, while the convolutional approach here described (let's call it ECL, for ellipse-center likelihood) takes about 2.5 seconds per image.

\begin{figure}[p]
\centering
\vspace{1cm}
\includegraphics[width=0.8\linewidth]{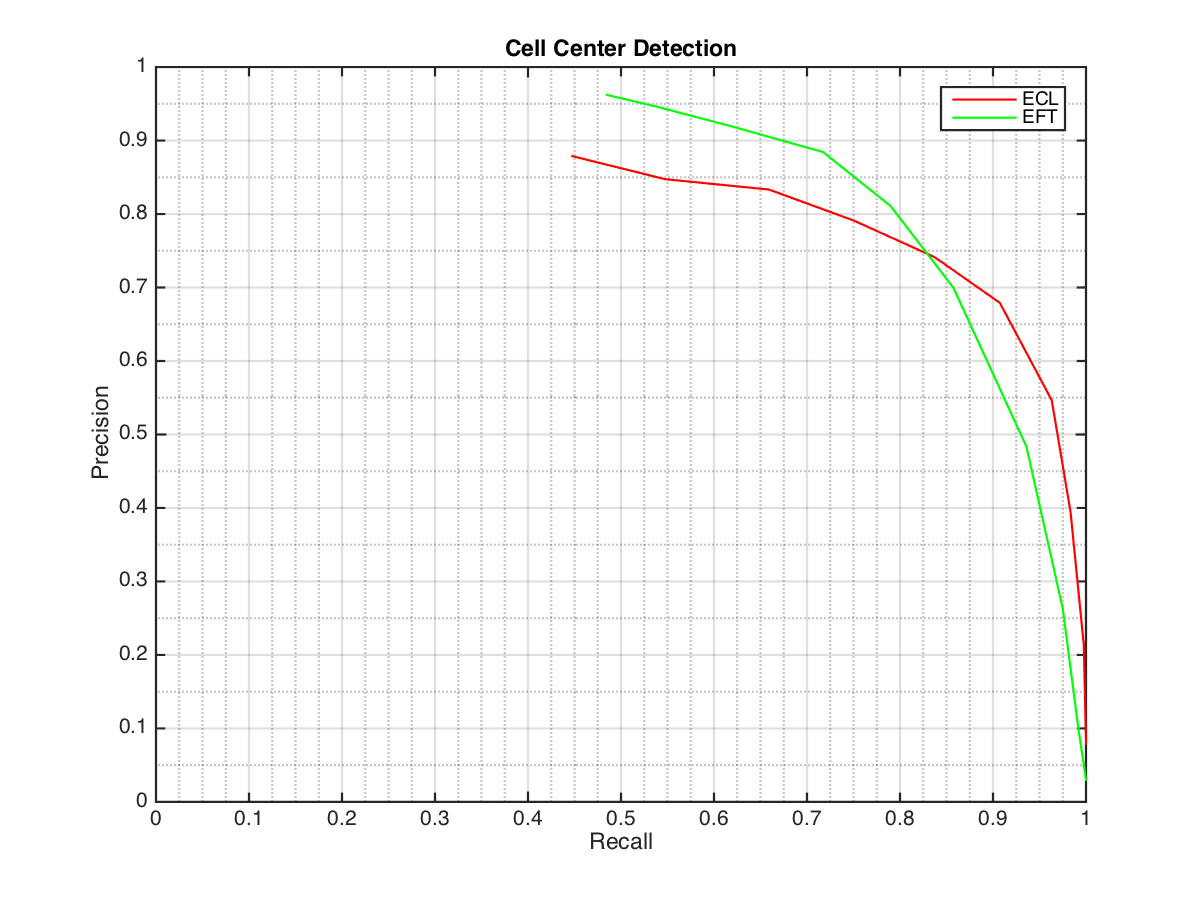}
\caption{Performance of the feature-centered EFT algorithm versus the convolutional algorithm in locating centers of cells in a dataset of 421 images from \cite{Cicconet2014ICIP1}. While there is no clear winner in terms of performance, the situation is different in terms of computation time: in average 2.5 seconds per image for convolutional, versus 87.5 seconds per image for EFT.}
\label{fig:ellipsedetection}
\end{figure}

Figure~\ref{fig:ellipsedetection} shows the performance comparison in terms of precision/recall curves. The curves were generated by varying the threshold above which local maxima in the accumulator spaces were detected (from $0.1$ to $1$). As per Figure~\ref{fig:ellipsedetection}, the performances are similar, with different methods being better in different segments of the recall range. Thus, depending on the application and circumstances, the ellipse-center likelihood derived from the convolutional symmetry method might be the best choice, especially if time is a main concern. 

\section{Conclusion}

We introduced a new approach to the problem of mirror symmetry detection in 2D, by posing the search for parameters from the point of view of the parameters themselves, instead of from the point of view of image-features, as has been done in the past.

Of course, there are trade-offs to consider, and our method might not be the best choice when the number of image features is or can be significantly reduced.
On the other hand, if properties of the shape are known and the search space of parameters can be reduced, than our method might perform better. We showed one such case in the problem of detecting near circular shapes (the same principle applies to symmetry line/segment detection).

Finally, there is work to be done in terms of improving the location of end points, given a symmetry line. One possibility would be to intersect the symmetry line with the area given by an object-proposal algorithm.

Our current research is focusing on how to embed this type of geometric prior in convolutional neural
networks.

\bibliographystyle{plain}
\bibliography{refs}

\end{document}